# Towards Multi-Object Nonprehensile Transportation via Shared Teleoperation: A Framework Based on Virtual Object Model Predictive Control

Xinyang FAN[a], Zhaoyang CHEN[a], Shu XIN[a], Yi REN[b], Zainan JIANG[a], Fenglei NI(✉)[a], Hong LIU[a]

[a] State Key Laboratory of Robotics and System, Harbin Institute of Technology, Harbin 150001, China.

[b] Advanced Manufacturing Laboratory, Huawei Technologies, Shenzhen 518129, China.

**ABSTRACT** Multi-object nonprehensile transportation in teleoperated robotic systems poses a dual control challenge: real-time trajectory tracking and simultaneous tray orientation control to satisfy object dynamic constraints. Existing approaches face limitations, including difficulty satisfying trajectory state constraints, excessive model dependency, inadequate adaptability to multi-object scenarios, and a lack of robust mechanisms for handling uncertain object parameters. To address these limitations, this work proposes a novel shared teleoperation framework for multi-object nonprehensile transportation, which enables shared control between human operators and the robotic system for object positioning; meanwhile, the robot autonomously controls object orientation to satisfy task constraints. The primary contributions are threefold: First, a theoretical analysis of dynamic constraints is developed, incorporating object position, inertial parameters, quantity, friction coefficients, and motion states. Furthermore, a virtual object (VO)-based dynamic constraint processing method is proposed for the first time, enabling simplified dynamic constraints to be directly utilized for trajectory planning. Second, a model predictive control (MPC)-based trajectory smoothing algorithm with real-time dynamic constraint enforcement is designed, enabling dynamic coordination between user input tracking and orientation control. Third, simulation and experimental validation confirm that the proposed method successfully ensures dynamic constraints for all objects and achieves stable manipulation of nine different objects at accelerations up to 2.4 m/s². Compared to the baseline method, the approach achieves a 72.45% reduction in sliding distance and maintains a zero tip-over rate (compared to 13.9% for the baseline). These results demonstrate enhanced adaptability to multi-object parameters and robust performance in complex nonprehensile transportation scenarios.

## 1 Introduction

Fully autonomous robots remain insufficient for meeting real-world task requirements, making teleoperation control a critical means for accomplishing complex robotic operations [1,2]. This control paradigm leverages human cognitive capabilities in conjunction with robotic precision and speed to enhance task execution performance. In this context, shared teleoperation control can reduce operator workload and improve task performance by combining operator control with autonomous robot control

E-mail: flni@hit.edu.cn
Tel: 18604513569

[3]. This approach proves particularly advantageous for complex manipulation tasks and multi-degree-of-freedom robotic systems [4].

This study addresses the teleoperation control challenges associated with robotic nonprehensile transportation tasks. Nonprehensile multi-object transportation represents a critical application scenario in industrial automation and service robotics [5]. The unilateral constraints inherent in nonprehensile transport—where objects are typically carried on a tray—impose a dual control objective on operators. On one hand, they must precisely control the robot end-effector's position to achieve the transport goal. On the other hand, they must simultaneously control its attitude. The latter is critical for satisfying the objects' friction cone (FC) and Zero-Moment Point (ZMP) constraints, which prevents the objects from sliding, toppling, or losing contact.

In recent years, learning-based methods have garnered significant attention in robotics manipulation tasks [6–11]. However, to date, no learning-based methods have been specifically applied to nonprehensile transportation. Methods based on Imitation Learning (IL) execute tasks by learning from human demonstrations [12–14]. Yet, human operators, when performing nonprehensile transportation, tend to adopt extremely slow acceleration/deceleration strategies during the start and end phases. This is done by reducing motion acceleration to prevent excessive tray tilting. Although this conservative strategy guarantees object stability, it undoubtedly reduces operational flexibility and efficiency, which is clearly a suboptimal motion trajectory. Similarly, while methods based on Reinforcement Learning (RL) through physics simulation have shown promising prospects and notable performance in many tasks [15,16], no relevant methods have yet been applied to nonprehensile transportation.

Learning-based approaches exhibit certain limitations when handling tasks that involve high-precision physical interactions (e.g., contact, friction) or high dynamics (requiring high-frequency real-time state feedback) [17]. They often require integration with traditional control methods to compensate for the deficiencies of pure learning-based approaches in real-time performance, robustness, and safety [18–20]. Consequently, given the stringent requirements of nonprehensile manipulation tasks—namely, precise constraint satisfaction, trajectory smoothness, and real-time capability—existing research has predominantly adopted model-based approaches to address this problem.

Extensive model-based motion planning studies for nonprehensile transportation have established foundational frameworks for this domain. Algorithm design typically focuses on two main aspects: efficiency and robustness on one hand, and stability with system constraints on the other hand. Regarding efficiency and robustness, the emphasis is on time-optimality and uncertainty tolerance. To solve time-optimal trajectories, previous studies have integrated robot dynamic parameter identification, end-path parameterization, and multiple-shot methods while considering friction, speed, and torque constraints [21,22]. Meanwhile, other work has utilized direct transcription and convex optimization techniques to solve shortest-time control for nonprehensile transportation problems [23]. Approaches to enhance robustness include building MPC controllers based on minimum friction coefficients to improve adaptation to friction and object inertia uncertainty [24,25], calculating inertial forces and reorienting objects to avoid sliding during high-speed operations [26], and applying quadratic programming (QP) for real-time control input computation with adaptive tuning mechanisms for tray orientation to minimize sliding [27]. Regarding stability and system constraints, the focus is on safe task completion under various perturbations and robot limitations. Some researchers have modeled objects as three-dimensional (3D) linear inverted pendulums and implemented dynamic balance control on humanoid robots using ZMP principles and force sensors [28]. For legged robots and mobile-manipulator systems, whole-body control architectures have been designed to prevent object sliding and maintain robot balance [29]. In dual-arm manipulation scenarios, time-optimal path parameterization has been combined with MPC for trajectory planning and impedance controllers to ensure accurate contact forces and avoid slippage [30]. Additionally, methods have been developed to explicitly incorporate joint position, velocity, and torque constraints as system constraints within MPC frameworks such as MPNSM [31].

Teleoperation tasks involve trajectories generated from real-time user inputs, which frequently contain discontinuous or impulsive commands. Therefore, controllers must perform online trajectory smoothing while maintaining task constraints. Simultaneously, nonprehensile transportation demands that the tangential forces exerted on the objects be minimized to prevent slipping and maintain contact stability. This dual objective—achieving a smooth trajectory while satisfying dynamic constraints—is strongly analogous to the problem of slosh-free liquid transport. Therefore, control strategies from liquid transport provide valuable insights for nonprehensile manipulation research. Point-mass spherical pendulum models with linearized dynamics have enabled MPC-based sloshing suppression [32,33], while damped pendulum models have facilitated input shaping approaches [34,35]. A different strategy has been inspired by quadrotor control principles to achieve sloshing-free transportation [36]. Additionally, input shaping and filtering techniques have been widely adopted to enhance trajectory continuity and dynamic characteristics in teleoperation systems [37–46].

Given the constraint characteristics inherent in nonprehensile transportation, shared control methods have emerged as effective solutions. These methods enable shared control between human operators and robotic systems for object positioning, while the robot autonomously controls object orientation to satisfy task constraints. Selvaggio et al. [47] conducted the earliest research on shared teleoperation methods for nonprehensile object transportation. Their approach enabled nonprehensile transportation of various objects on a tray by introducing virtual moments that correlate with friction forces to control object orientation, while ensuring object dynamics constraints via QP.

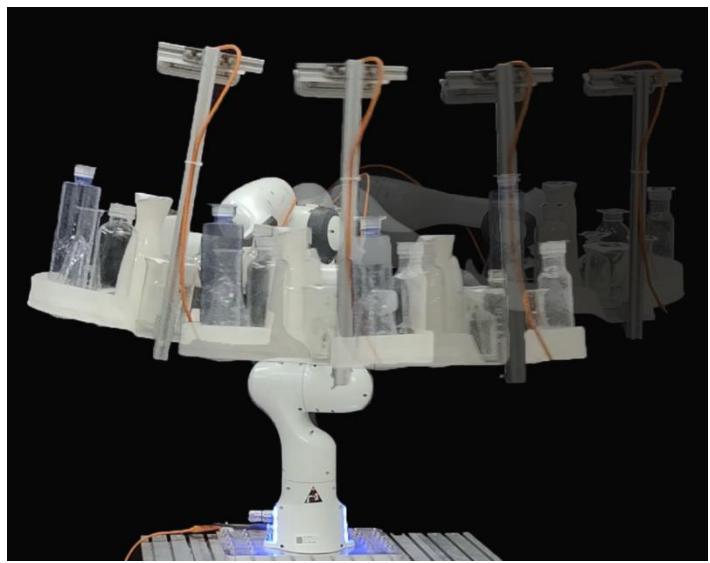

**Fig. 1** Multi-Object nonprehensile transportation through VO-MPC architecture. A video of our experiments is available at https://youtu.be/NsWn40RT-I0.

However, existing methods suffer from the following limitations:

First, nonprehensile tasks impose explicit state constraints on robot trajectories, including workspace limitations, velocity and acceleration bounds, as well as task-related requirements such as FC and ZMP constraints. Current real-time trajectory smoothing methods typically modify the frequency-domain characteristics of the trajectory through techniques such as filtering and input shaping. However, a critical drawback is that the trajectory's state constraints remain difficult to satisfy.

Second, both model-based planning and teleoperation algorithms require accurate object dynamic models and position information to maintain stability. Each approach requires dynamic modeling or model simplification of the transported objects and the enforcement of contact force constraints. In practice, transported objects vary significantly in shape, size, and placement, making precise dynamic models and positions unrealistic assumptions.

Third, existing algorithms manifest poor scalability to multi-object scenarios, limiting transportation efficiency. Most approaches target single objects, with few addressing multi-object transportation. The latter require complete prior knowledge of each object's position, shape, and inertial properties, which severely constrains algorithm generality and robustness.

Fourth, existing algorithms lack effective mechanisms for handling dynamic constraints when object parameters such as number, position, inertial properties, and friction coefficients are uncertain. While some algorithms show adaptability to object geometry and placement, rigorous theoretical proof of this capability is not provided.

To address these limitations, this work proposes a robust shared teleoperation control framework for multi-object nonprehensile transportation (as illustrated in Fig. 1). The contributions include:

A theoretical analysis of the relationships between object parameters (inertial properties, position, quantity, motion state) and dynamic constraints is conducted. Subsequently, a novel VO-based dynamic constraint processing method is proposed for the first time, which yields simplified dynamic constraints that can be directly applied to trajectory planning for nonprehensile transportation tasks.

An MPC-based trajectory smoothing method is developed for real-time teleoperation. The method automatically generates smooth object trajectories from user position inputs while enforcing nonprehensile transportation constraints, achieving shared control for object positions and autonomous control for object orientations.

Both simulation and experimental validation confirm that the proposed method successfully ensures dynamic constraints for all objects while achieving smooth completion of multi-object nonprehensile transportation tasks with high trajectory tracking accuracy. Furthermore, the method exhibits robust performance across varying object parameters, including object number, inertial properties, positions, and friction coefficients.

To present the proposed approach and its validation, the remainder of this paper is organized as follows: Section 2 introduces a VO-based approach for processing multi-object dynamic constraints. Section 3 details the teleoperation control framework. Section 4 presents simulation experiments, while Section 5 provides hardware experiments. Section 6 concludes the paper.

**2 Virtual object-based constraint processing for multi-object dynamics**

Nonprehensile transportation achieves efficient object handling through unilateral constraints imposed on transported objects. This method relies on contact forces between objects and transport surfaces, combined with object inertial forces, to maintain stability during motion. The relationship between object motion states and force conditions depends on multiple factors including object positions on transport surfaces, inertial properties, geometric shapes, and contact surface friction coefficients. Dynamic constraints for transported objects can be derived through *Newton-Euler* equations based on current motion states and the aforementioned parameters. However, precise dynamic constraint formulations involve numerous coupling terms and nonlinearities, which prevent closed-form expressions and thereby create challenges for real-time constraint enforcement.

In practical applications, different factors exhibit varying degrees of influence on system constraints. Therefore, minor contributing terms can be reasonably neglected to improve computational efficiency with negligible effect on accuracy, a practice well-established in engineering applications [48,49]. Therefore, this section aims to derive simplified yet effective constraints applicable to trajectory planning through systematic constraint analysis and approximation, while ensuring robustness to parameter variations.

**Problem Statement:** Given the approximate ranges of object geometric and inertial parameters, as well as the tray dimensions, and subject to the robotic arm's performance limitations, this research aims to determine the motion state constraints that must be satisfied during teleoperation-based nonprehensile transportation to ensure the objects neither slide nor topple.

This problem is inherently coupled with object dynamic models. Regardless of the approach employed, achieving completely model-free control is unattainable. The objective is to minimize dependency on model parameters while maximizing method robustness to parameter variations and uncertainties. Therefore, for the theoretical analysis presented in this section, we initially assume that object positions, geometric properties, inertial parameters, and surface friction coefficients are known a priori. The assumption will be relaxed in subsequent sections to address practical uncertainties.

**2.1 Single Object**

Single object nonprehensile transportation represents the simplest case. First, consider constraints for preventing sliding. Objects move under tray control, where the greater the friction force exerted on the object, the higher the likelihood of sliding. To prevent object sliding, the friction force must be minimized. For any desired acceleration, there exists a desired tray orientation where the friction force exerted on the object becomes zero. In this orientation, the combined effect of gravity and normal support forces provides the entire desired object acceleration, as shown in Fig. 2 (where $F_S, F_g, \ddot{x}_d, g \in \mathbb{R}^3$ are the support force, gravity, desired object acceleration, and gravitational acceleration, respectively). The desired tray orientation is expressed using axis-angle representation as $\Phi = \alpha \hat{n}$, where $\hat{n}$ is the unit rotation axis ($\hat{n} \in \mathbb{R}^3, \|\hat{n}\| = 1$), and $\alpha$ is the rotation angle around $\hat{n}$:

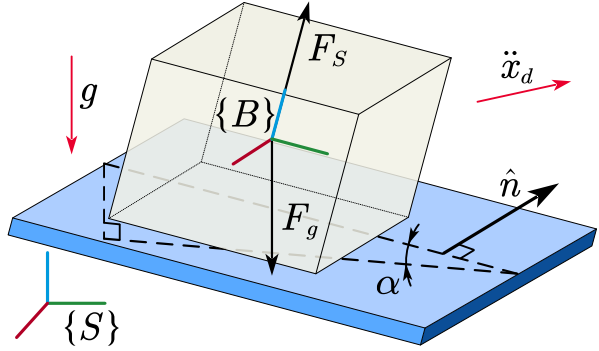

**Fig. 2** Friction-free orientation illustration.

$$\hat{n} = \frac{\ddot{x}_d \times g}{\|\ddot{x}_d \times g\|},$$
$$\alpha = \langle g, g - \ddot{x}_d \rangle = \arccos\left(\frac{g^T(g - \ddot{x}_d)}{\|g\|\|g - \ddot{x}_d\|}\right). \quad (1)$$

Theoretically, if object orientation and acceleration satisfy this relationship at all times, the object will not be subject to friction forces. However, from a rotational dynamics perspective, object orientation trajectories must satisfy at least $C^2$ continuity to prevent objects from impacts during rotation. From Eq. (1), object orientation satisfying $C^2$ continuity requires object acceleration to satisfy $C^2$ continuity, which means the object position must satisfy $C^4$ continuity [50]. Under this premise, when object O's center of mass (CoM) coordinate frame is employed as the tray's rotational center for orientation calculation, the friction force applied to the object can be maintained at zero.

Further consider the ZMP constraint of the object. The moment applied to the object has an upper limit, namely:

$$T = I\dot{\Omega} + \Omega \times I\Omega,$$
$$\|T\| \leqslant \|F_c\| \cdot \frac{d}{2},$$

where $T \in \mathbb{R}^3$ is the moment applied to the object; $I \in \mathbb{R}^{3\times 3}$ is the rigid body inertia tensor in CoM coordinate frames; $\Omega \in \mathbb{R}^3$ is object angular velocity; $\dot{\Omega} \in \mathbb{R}^3$ is angular acceleration; $F_c \in \mathbb{R}^3$ is contact forces; and $d$ is the diameter of the inscribed circle of the contact surface.

### 2.2 Two Objects

Assume that there are two objects on the tray, denoted as A and O, respectively. The algorithm performs trajectory planning with reference to object O, ensuring that the position trajectory of O is $C^4$ continuous, while using O's CoM coordinate frame as the rotational center for computing the tray orientation. Objects A and O share the same orientation, angular velocity, and angular acceleration; however, their linear velocities and accelerations differ. The acceleration of object A is given by:

$$\dot{V}_A = \dot{V}_o + \dot{\Omega}_o \times P_A + \Omega_o \times (\Omega_o \times P_A),$$

where $\Omega_o, \dot{\Omega}_o \in \mathbb{R}^3$ denote the angular velocity and angular acceleration of object O, respectively, and $\dot{V}_A, \dot{V}_o \in \mathbb{R}^3$ represent the linear accelerations of objects A and O, respectively. $P_A \in \mathbb{R}^3$ is the vector from the CoM frame of object O to that of object A, representing the position of object A relative to object O.

#### 2.2.1 Friction cone constraint

The dynamic equation of object A is as follows:

$$F_A = M_A \dot{V}_A = M_A \dot{V}_o + M_A (\dot{\Omega}_o \times P_A) + M_A \Omega_o \times (\Omega_o \times P_A),$$

where $F_A \in \mathbb{R}^3$ is the resultant force exerted on object A; $M_A \in \mathbb{R}^{3\times 3}$ is the mass matrix of object A. The contact force $F_{c,A}$ exerted on object A can be expressed as:

$$F_{c,A} = M_A (\dot{V}_o - g) + M_A (\dot{\Omega}_o \times P_A) + M_A \Omega_o \times (\Omega_o \times P_A). \tag{2}$$

According to the Eq. (1), $M_A (\dot{V}_o - g)$ is perpendicular to the contact surface, so the friction force exerted on object A comes from $M_A (\dot{\Omega}_o \times P_A) + M_A \Omega_o \times (\Omega_o \times P_A)$.

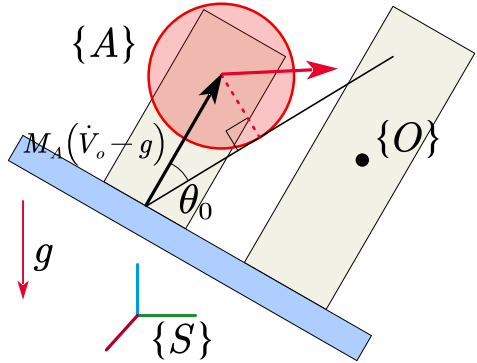

**Fig. 3** Schematic diagram of the FC constraint and force analysis of object A.

The force diagram for object A is illustrated in Fig. 3. Here, $g \in \mathbb{R}^3$ denotes the gravitational acceleration. Let $\theta_0 = \arctan \mu$, where $\mu$ is the coefficient of friction between object A and the tray. In Fig. 3, the black arrow represents $M_A (\dot{V}_o - g)$, while the red arrow represents $M_A (\dot{\Omega}_o \times P_A) + M_A \Omega_o \times (\Omega_o \times P_A)$. Therefore,

$$\left\| M_A (\dot{\Omega}_o \times P_A) + M_A \Omega_o \times (\Omega_o \times P_A) \right\| \leqslant M_A \left\| \dot{V}_o - g \right\| \sin \theta_0,$$

i.e., the term $M_A (\dot{\Omega}_o \times P_A) + M_A \Omega_o \times (\Omega_o \times P_A)$ lying within the region indicated by the red circle in Fig. 3 can serve as a sufficient condition for ensuring that object A satisfies the friction cone constraint.

Experimental observations indicate that under the maximum tray acceleration (2.4 m/s²) and maximum tilt angle (13.7°) employed in this study, the contribution of the term $M_A \Omega_o \times (\Omega_o \times P_A)$ is negligible ( $\|M_A \Omega_o \times (\Omega_o \times P_A)\| < 0.05 \|M_A (\dot{\Omega}_o \times P_A) + M_A \Omega_o \times (\Omega_o \times P_A)\|$ ), and its influence on the friction cone constraint can be ignored. Therefore, in the subsequent analysis, this minor term can be omitted, yielding:

$$F_{c,A} \approx M_A (\dot{V}_o - g) + M_A (\dot{\Omega}_o \times P_A). \qquad (3)$$

Since $\|M_A (\dot{\Omega}_o \times P_A)\| \leqslant M_A \|\dot{\Omega}_o\| \|P_A\|$, a sufficient condition for $F_{c,A}$ to always satisfy the friction cone constraint can be expressed as:

$$\|\dot{\Omega}_o\| \leqslant \frac{\|\dot{V}_o - g\|}{\|P_A\|} \sin\theta_0, \qquad (4)$$

Given that $\sin\theta_0 \leqslant 1$, satisfying the constraint in Eq. (4) also ensures that the object remains in contact with the tray at all times.

### 2.2.2 Zero-Moment Point constraint

Assume that object A is a cuboid with a square base of side length $d$ and height $h$, as shown in Fig. 4.

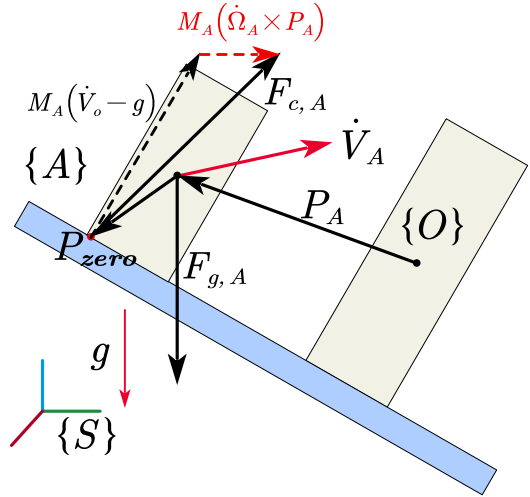

**Fig. 4** Schematic diagram of ZMP constraint and force analysis of object A.

The *Euler equation* of object A is as follows:

$$T_A = I_A \dot{\Omega}_A + \Omega_A \times I_A \Omega_A, \qquad (5)$$

$$T_A = P_{zero} \times F_{c,A}, \qquad (6)$$

where $T_A \in \mathbb{R}^3$ is the moment applied to object A; $\Omega_A, \dot{\Omega}_A \in \mathbb{R}^3$ are the angular velocity and angular acceleration of object A, respectively; $\Omega_A = \Omega_o$, $\dot{\Omega}_A = \dot{\Omega}_o$; $I_A \in \mathbb{R}^{3\times 3}$ is the inertia matrix of object A; $P_{zero} \in \mathbb{R}^3$ is the ZMP of object A; and $F_{c,A} \in \mathbb{R}^3$ is the contact force applied to object A.

Experimental observations indicate that, under the maximum tray acceleration used in this study (2.4 m/s²) and maximum tilt angle (13.7°), the contribution of the term $\Omega_A \times I_A \Omega_A$ is negligible ( $\|\Omega_A \times I_A \Omega_A\| < 0.03 \|I_A \dot{\Omega}_A + \Omega_A \times I_A \Omega_A\|$ ), and its effect on the ZMP constraint can be ignored. Therefore, Eq. (5) can be simplified as:

$$T_A \approx I_A \dot{\Omega}_A. \qquad (7)$$

Equation (7) represents the torque required for the motion, while Eq. (6) represents the torque provided by the tray. During motion, these torques always remain aligned in the same direction. Consequently, for object A to satisfy the ZMP constraint, the required torque must not exceed the maximum torque provided by the tray. Substituting Eq. (3) yields:

$$\|I_A \dot{\Omega}_A\| \leqslant \|P_{zero} \times M_A (\dot{V}_o - g) + P_{zero} \times M_A (\dot{\Omega}_A \times P_A)\|_{max}. \tag{8}$$

When the object reaches its maximum angular acceleration, it is on the verge of tipping over, which implies:

$$\|P_{zero} \times M_A (\dot{V}_o - g)\|_{max} \geqslant M_A \|\dot{V}_o - g\| \frac{d}{2}. \tag{9}$$

For computational convenience and to ensure sufficiency, the following relationship is adopted:

$$\|P_{zero} \times M_A (\dot{V}_o - g + \dot{\Omega}_A \times P_A)\|_{max} \geqslant M_A \|\dot{V}_o - g\| \frac{d}{2} - \|P_{zero} \times M_A (\dot{\Omega}_A \times P_A)\|_{max}.$$

Since $\|P_{zero} \times M_A (\dot{\Omega}_A \times P_A)\|_{max} \leqslant \frac{1}{2} M_A \|\dot{\Omega}_A\| \|P_A\| \sqrt{d^2 + h^2}$ and $\|I_A \dot{\Omega}_A\| \leqslant \|\dot{\Omega}_A\| J_{max}$, where $J_{max}$ is the maximum rotational inertia of object A, the sufficient condition for object A to satisfy the ZMP constraint is as follows:

$$\|\dot{\Omega}_o\| = \|\dot{\Omega}_A\| < \frac{\|\dot{V}_o - g\| d}{2 M_A^{-1} J_{max} + \|P_A\| \sqrt{d^2 + h^2}}. \tag{10}$$

In summary, considering both the friction cone and ZMP constraints, the angular acceleration of object A should be constrained by the minimum of Eqs. (4) and (10), i.e.,

$$\|\dot{\Omega}_A\| \leqslant \min \left\{ \frac{\|\dot{V}_o - g\|}{\|P_A\|} \sin \theta_0, \frac{\|\dot{V}_o - g\| d}{2 J_{max} M_A^{-1} + \|P_A\| \sqrt{d^2 + h^2}} \right\} \tag{11}$$

**2.3 Multiple Objects**

The analysis of dynamical constraints for multiple objects can be generalized from the above two-object framework. Important conclusions emerge from Eqs. (4) and (10):

**Sliding behavior**: The tendency for object A to slide depends on the distance between the CoMs of objects A and O, and the friction coefficient. Larger inter-CoM distances and smaller friction coefficients increase the likelihood of sliding.

**Tipping behavior**: The tendency for object A to tip over depends on its relative mass distribution ($M_A^{-1} J_{max}$) and the distance between the CoMs of objects A and O. Larger inter-CoM distances and higher values of the mass distribution increase the likelihood of tipping. This criterion is independent of the absolute values of mass and inertia, depending only on their ratio.

**Note:** Since both the torque exerted on the object and the torque required for motion are proportional to mass, the mass does not affect the rotational constraints of the object.

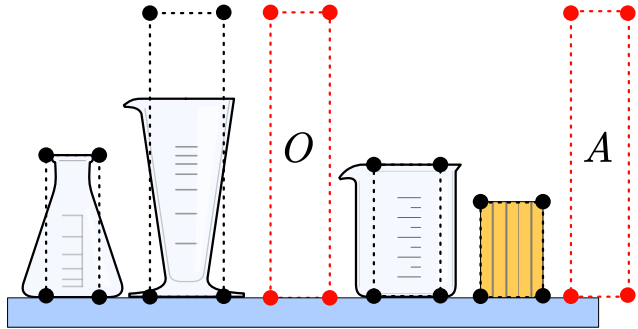

**Fig. 5** Multi-object dynamic constraint analysis. Black dashed boxes denote the original object

approximations, while the two red dashed boxes denote the reference object O and offset object A. All dashed boxes indicate objects with mass concentrated at their vertices. The blue rectangle at the bottom represents the tray.

When multiple objects are carried on a tray, the following procedure is implemented:

(1) Object Approximation Setup.

Each object is approximated by a cuboid with a square base. As illustrated in Fig. 5, objects such as a conical flask, a measuring cup, a beaker, and a rectangular wooden block are each approximated by a cuboid (shown as black dashed boxes). The base of each approximating cuboid corresponds to the largest inscribed square of the object's contact surface with the tray, with a side length of $d_i$, and its height $h_i$ is identical to that of the actual object. For objects with higher centers of gravity, the approximating height can be conservatively increased (e.g., the measuring cup is approximated using a cuboid with height 1.5 times its actual height to ensure more restrictive constraints).

(2) Mass and Inertia Configuration.

Each approximating cuboid is assigned a mass of 1 kg, concentrated at its 8 vertices. This configuration maximizes the rotational inertia for the given mass and geometry, making the approximated object more susceptible to tipping than the original object. The friction coefficient $\mu_i$ for each approximating object is estimated based on the material properties of the corresponding original object and prior knowledge.

(3) Reference and Offset Object Definition.

Two special virtual objects are introduced for trajectory planning and dynamic constraint analysis:

**Virtual reference object O**: Designed for trajectory planning, this object is located at the geometric center of the tray and serves as the reference to ensure $C^4$ position continuity. The CoM coordinate frame of object O is used as the rotational center for tray orientation control. Object O has the same shape and dimensions as the offset object A.

**Virtual offset object A**: Designed for dynamic constraint analysis, this cuboid has a square base with 1 kg mass concentrated at its 8 vertices. Its dimensions are determined by $d = \min\{d_i\}$ for the base side length and $h = \max\{h_i\}$ for the height. The inertia matrix of the offset object is $I_{\max} = \mathrm{diag}\left(\frac{1}{4}(d^2+h^2),\ \frac{1}{4}(d^2+h^2),\ \frac{1}{2}d^2\right)$. Object A is positioned to maximize the distance from object O, specifically at distance $\|P_A\| = \sqrt{R^2+(0.5h)^2}$ from the reference object, where $R$ is the maximum radius of the actual tray. The friction coefficient $\mu$ between object A and the tray is defined as $\mu = \min\{\mu_i\}$.

This configuration renders the offset object more susceptible to sliding and tipping compared to all objects on the tray. Consequently, regardless of the number of objects placed on the tray, the offset object consistently exhibits the highest instability and becomes the first to violate the constraints. Therefore, the dynamic constraints for multiple objects are transformed into constraints for a single offset object.

The dynamic constraints based on the offset object are inherently robust to the number and positions of the objects. This is because the parameters of the offset object ($d, h, \|P_A\|, \mu$) are independent of the number and positions of the objects on the tray. By contrast, the robustness of the constraints to object shape, inertial parameters, and friction coefficients is related to the estimation of object parameters, since the parameters of the offset object must be determined according to the objects on the tray. Although it is difficult to accurately estimate the shape, size, inertia, and friction coefficient of each object, the algorithm only requires the extrema of these parameters. Even when the estimation is inaccurate, more conservative constraints can still be obtained in practice by appropriately reducing

$\mu$ and increasing $h$ (or decreasing $d$). Therefore, although the constraints are influenced by parameter estimation, they are not completely dependent on it.

Some existing methods rely on the real-time dynamic parameters of the objects to guarantee constraint satisfaction, whereas the proposed method does not require the parameters of the offset object to be updated in real time, as long as its *most critical* property is ensured. Therefore, once the offset object is determined, the process by which the controller enforces the constraints is no longer related to the objects on the tray. Although this problem is theoretically impossible to solve in a completely model-free manner, the introduction of the offset object greatly reduces the dependence of constraint satisfaction on object model parameters.

### 3 Teleoperation control framework

When controlling robots for nonprehensile transportation tasks through teleoperation, operators input desired object positions online via master devices. Since object orientation has no specific requirements beyond satisfying constraints, the operator does not need to control it. Therefore, the controller must smooth user inputs while satisfying the dynamics constraints presented in the previous section and autonomously controlling object orientation according to Eq. (1). The teleoperation control framework is illustrated in Fig. 6.

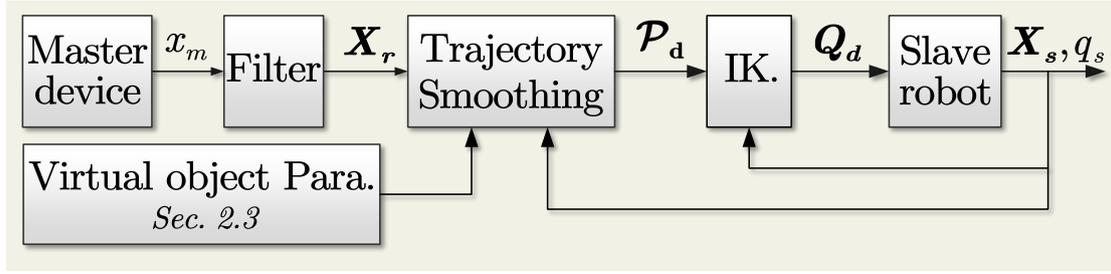

**Fig. 6** Teleoperation control framework.

Here, $x_m$ denotes the operator's hand position acquired through the master device, which also serves as the input trajectory for the controller. The stochasticity of user operation and physiological tremors introduce noise into $x_m$. After filtering, the reference position $x_r$ is obtained, with $x_m, x_r \in \mathbb{R}^3$. $\boldsymbol{X}_r$ represents the input reference trajectory for the trajectory smoother, $\boldsymbol{X}_r = \left[x_r^\mathrm{T}, \dot{x}_r^\mathrm{T}, \ddot{x}_r^\mathrm{T}\right]^\mathrm{T}$, where $\dot{x}_r$ and $\ddot{x}_r$ are the reference velocity and acceleration, respectively, obtained through first- and second-order differentiation of $x_r$. $\mathcal{P}_d$ denotes the output of the trajectory smoother, representing the desired end-effector position and orientation of the reference object, $\mathcal{P}_d = \left[x_d^\mathrm{T}, \Phi_d^\mathrm{T}\right]^\mathrm{T}$, where $x_d \in \mathbb{R}^3$ is the desired position and $\Phi_d \in \mathbb{R}^3$ is the desired orientation.

The output of the inverse kinematics $\boldsymbol{Q}_d$ represents the desired joint trajectory of the robot, $\boldsymbol{Q}_d = \left[q_d^\mathrm{T}, \dot{q}_d^\mathrm{T}\right]^\mathrm{T}$, where $q_d, \dot{q}_d \in \mathbb{R}^n$ denote the desired joint positions and velocities, respectively, and $n$ is the number of robot joints. $\boldsymbol{X}_s$ denotes the actual state feedback from the slave robot, $\boldsymbol{X}_s = \left[x_s^\mathrm{T}, \dot{x}_s^\mathrm{T}, \ddot{x}_s^\mathrm{T}\right]^\mathrm{T}$, where $x_s, \dot{x}_s, \ddot{x}_s \in \mathbb{R}^3$ are the robot's actual end-effector position, velocity, and acceleration. $q_s$ represents the actual joint angles of the robot, $q_s = \left[q_{s,1}, q_{s,2}...q_{s,n}\right]^\mathrm{T}$.

For redundant manipulators ($n > 6$), the inverse kinematics algorithm [51] enables manipulator control while ensuring object convergence to desired positions and orientations, which is presented as follows:

$$\begin{cases} \dot{q}_d = J^\dagger\left(\mathcal{V}_d + K_e\left(\mathcal{P}_d - Jq_s\right)\right) + \left(\mathbf{I}_n - J^\dagger J\right)\dot{q}_0, \\ K_e = \begin{bmatrix} K_{e,x} & \mathbf{0}_3 \\ \mathbf{0}_3 & K_{e,\varphi} \end{bmatrix}, \dot{q}_0 = K_H \nabla H(q_s), \\ H(q) = \sum_{i=1}^{n} \frac{(\bar{q}_i - \underline{q}_i)^2}{(\bar{q}_i - q_{s,i})(q_{s,i} - \underline{q}_i)}, \end{cases}$$

where $\mathcal{V}_d$ is desired end-effector velocities and angular velocities, which is obtained through differentiation of $\mathcal{P}_d$; $K_{e,x}, K_{e,\varphi} \in \mathbb{R}^{3\times 3}$ are the position and the orientation error weight matrices, respectively; $J$ and $J^\dagger$ are the manipulator Jacobian matrix and its pseudo-inverse; $\dot{q}_0 \in \mathbb{R}^n$ is manipulator null space motion for low-priority joint limit avoidance subtasks; $K_H \in \mathbb{R}^{n\times n}$ is a positive definite diagonal matrix; $\bar{q}_i$ and $\underline{q}_i$ are the upper and the lower limits of the $i$-th joint, respectively.

### 3.1 MPC-Based Trajectory Smoothing

To facilitate constraint handling, an MPC-based trajectory smoother is designed to generate C⁴ continuous position trajectories in real time, as illustrated in Fig. 7. We consider a control system comprising five cascaded integrators for position control. The objective of the controller is to compute the optimal system input $u$ such that $X_s$ tracks $X_r$. In Fig. 7, $X_d$ denotes the desired state of the trajectory smoother, defined as $X_d = \left[x_d^T, \dot{x}_d^T, \ddot{x}_d^T, \dddot{x}_d^T, \ddddot{x}_d^T\right]^T$, where $x_d \in \mathbb{R}^3$ represents the desired position of the reference object. The system input $u \in \mathbb{R}^3$ corresponds to the fifth-order derivative of the desired position. The controller output position trajectory is coupled with orientation through Eq. (1), ultimately generating the desired position and orientation trajectory of the reference object, denoted as $\mathcal{P}_d$.

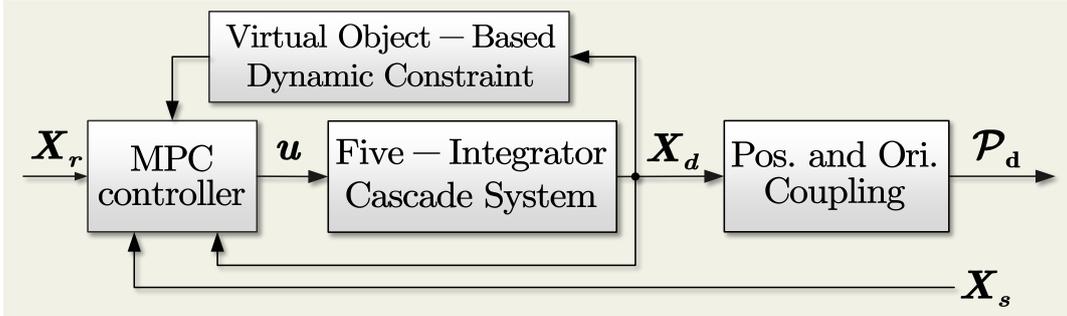

**Fig. 7** The control block diagram of the trajectory smoother.

The state equation of the integral system is as follows:

$$\begin{cases} \dot{X}_d = A_d X_d + B_d u, \\ A_d = \begin{bmatrix} \mathbf{0}_{12\times 3} & \mathbf{I}_{12} \\ \mathbf{0}_3 & \mathbf{0}_{3\times 12} \end{bmatrix}, \\ B_d = \begin{bmatrix} \mathbf{0}_{12\times 3} \\ \mathbf{I}_3 \end{bmatrix}, \end{cases}$$

To facilitate controller design, an augmented system is constructed using the desired state $X_d$ and actual state $X_s$. This augmented system is described as follows:

$$\begin{cases} \begin{bmatrix} \dot{X}_d \\ \dot{X}_s \end{bmatrix} = \begin{bmatrix} A_d & 0_9 \\ A_{s,d} & A_{s,s} \end{bmatrix} \begin{bmatrix} X_d \\ X_s \end{bmatrix} + \begin{bmatrix} B_d \\ 0_{9\times 3} \end{bmatrix} u, \\ A_{s,d} = \begin{bmatrix} K_{e,x} & I_3 & 0_3 & 0_3 & 0_3 \\ 0_3 & K_{e,x} & I_3 & 0_3 & 0_3 \\ 0_3 & 0_3 & K_{e,x} & I_3 & 0_3 \end{bmatrix}, \\ A_{s,s} = \begin{bmatrix} -K_{e,x} & 0_3 & 0_3 \\ 0_3 & -K_{e,x} & 0_3 \\ 0_3 & 0_3 & -K_{e,x} \end{bmatrix}. \end{cases}$$

The above system is discretized as follows using the zero-order hold method with a given sampling time step:

$$\begin{cases} \mathcal{X}(k+1) = \mathcal{A}\mathcal{X}(k) + \mathcal{B}u(k), \\ \mathcal{X}(k) = \begin{bmatrix} X_d^{\mathrm{T}}(k) & X_s^{\mathrm{T}}(k) \end{bmatrix}^{\mathrm{T}}. \end{cases}$$

The standard MPC expression for a prediction interval of $N$ is as follows:

$$\begin{cases} \min_u \sum_{k=0}^{N-1} \left( \| X_r(k+1) - X_s(k+1) \|_{W_x}^2 + \| u(k) \|_{W_u}^2 \right), \\ \text{s.t.:} \mathcal{X}(k+1) = \mathcal{A}\mathcal{X}(k) + \mathcal{B}u(k), \\ \overline{\dot{x}} \leqslant \dot{x}_d(k+1) \leqslant \underline{\dot{x}}, \overline{\dot{x}} \leqslant \dot{x}_s(k+1) \leqslant \underline{\dot{x}}, \\ \overline{\ddot{x}} \leqslant \ddot{x}_d(k+1) \leqslant \underline{\ddot{x}}, \overline{\ddot{x}} \leqslant \ddot{x}_s(k+1) \leqslant \underline{\ddot{x}}, \\ \overline{\dddot{x}} \leqslant \dddot{x}_d(k+1) \leqslant \underline{\dddot{x}}, \overline{\dddot{x}} \leqslant \dddot{x}_s(k+1) \leqslant \underline{\dddot{x}}, \\ \overline{u} \leqslant u(k) \leqslant \underline{u}, \overline{\ddddot{x}} \leqslant \ddddot{x}_d(k+1) \leqslant \underline{\ddddot{x}}, k \in \{0, \cdots, N-1\}. \end{cases} \quad (12)$$

Where $\|\cdot\|_W^2 = (\cdot)^{\mathrm{T}} W(\cdot)$, $W_x$ and $W_u$ are the state and the input weight matrices, respectively; the optimization term $\|u(k)\|_{W_u}^2$ prevents high-frequency switching of the control output between upper and lower limits; $\mathcal{X}(0)$ is the current system state; manipulators are subject to end-effector velocity, acceleration, and jerk constraints, which impose corresponding state constraints; and $\overline{u} \leqslant u(k) \leqslant \underline{u}$ ensures continuity of the fourth-order derivative of position.

$\overline{\dddot{x}} \leqslant \dddot{x}(k) \leqslant \underline{\dddot{x}}$ represents the dynamic constraints derived from the virtual object in Section 2 (Eq. (11)). However, Eq. (11) cannot be directly applied to the trajectory smoother and thus requires the following processing.

The second-order time derivative of Eq. (1) yields the angular acceleration of the tray: $\dot{\Omega}_o = \dot{\Omega}_A = \dfrac{d^2}{dt^2}(\alpha \hat{n})$. Substituting (1) into this expression gives:

$$\dot{\Omega}_A = \frac{\dddot{x}_d \times g}{\|\ddot{x}_d \times g\|}\alpha + 2\frac{\ddot{x}_d \times g}{\|\ddot{x}_d \times g\|}\dot{\alpha} - 2\frac{(\ddot{x}_d \times g)^{\mathrm{T}}(\dddot{x}_d \times g)(\ddot{x}_d \times g)}{\|\ddot{x}_d \times g\|^3}\alpha$$

$$+\frac{\ddot{x}_d \times g}{\|\ddot{x}_d \times g\|}\ddot{\alpha} - 2\frac{(\ddot{x}_d \times g)^{\mathrm{T}}(\dddot{x}_d \times g)(\ddot{x}_d \times g)}{\|\ddot{x}_d \times g\|^3}\dot{\alpha} - \frac{(\ddot{x}_d \times g)^{\mathrm{T}}(\dddot{x}_d \times g)(\ddot{x}_d \times g)}{\|\ddot{x}_d \times g\|^3}\alpha \quad (13)$$

$$-\frac{(\ddot{x}_d \times g)^{\mathrm{T}}(\dddot{x}_d \times g)(\ddot{x}_d \times g)}{\|\ddot{x}_d \times g\|^3}\alpha + 3\frac{\left((\ddot{x}_d \times g)^{\mathrm{T}}(\dddot{x}_d \times g)\right)^2(\ddot{x}_d \times g)}{\|\ddot{x}_d \times g\|^5}\alpha.$$

Eq. (13) can be simplified as follows:

$$\dot{\Omega}_A \approx \begin{cases} \dfrac{\dddot{x} \times g}{\|\ddot{x} \times g\|}\alpha, & \alpha \neq 0, \\ \dfrac{\dddot{x} \times g}{\|g\|\|\ddot{x} - g\|}, & \alpha = 0. \end{cases} \quad (14)$$

In conjunction with Eq. (11), the trajectory smoother computes online the constraints on the fourth-order derivative of position in Eq. (12) based on the current acceleration and orientation. The constraint of the fourth-order derivative of position is expressed as:

$$\|\dddot{x}_d\| \leqslant \begin{cases} \dfrac{1}{\alpha}\|\dot{\Omega}_A\|_{\max}\sqrt{\ddot{x}_{d,1}^2 + \ddot{x}_{d,2}^2}, & \alpha \neq 0, \\ \|\dot{\Omega}_A\|_{\max}|g_3 - \ddot{x}_{d,3}|, & \alpha = 0. \end{cases} \quad (15)$$

Where $\ddot{x}_d = \begin{bmatrix} \ddot{x}_{d,1} & \ddot{x}_{d,2} & \ddot{x}_{d,3} \end{bmatrix}^{\mathrm{T}}$, $g = \begin{bmatrix} 0 & 0 & g_3 \end{bmatrix}^{\mathrm{T}}$. A detailed derivation of Eqs. (13), (14), and (15) is provided in the Appendix A.

### 4 Simulation

Simulation validation was performed to verify the effectiveness of the multi-object constraint processing method and trajectory smoother. A predefined trajectory was input to the controller to avoid user input discrepancies. The theoretical performance was evaluated by checking constraint compliance and measuring object trajectories in the simulation environment.

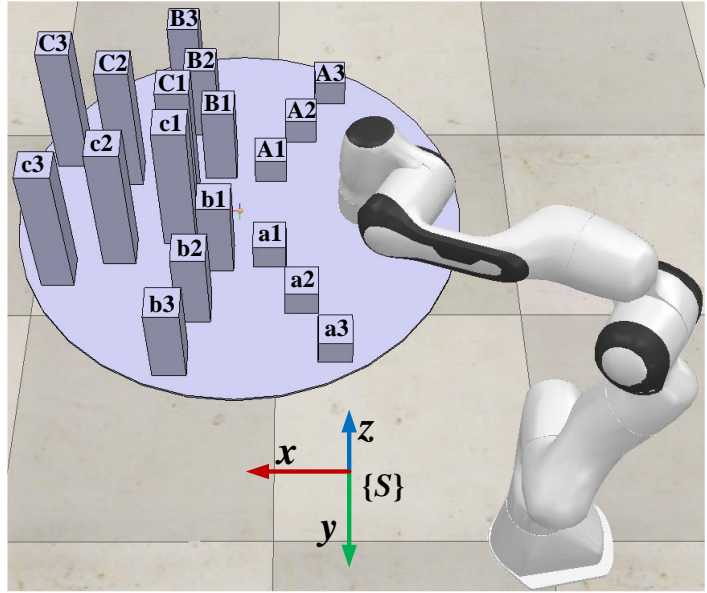

**Fig. 8** Simulation Configuration: A manipulator end-effector equipped with a tray transports 18 distinct objects. The system executes predefined trajectory inputs through manipulator controllers while detecting constraint violations related to ZMP and FC conditions.

### 4.1 Simulation Setup

The simulation was conducted in the CoppeliaSim environment using a Franka Emika Panda robot equipped with a tray mounted on the end-effector flange. The experimental setup involved 18 rectangular blocks positioned on the tray, with their locations and numbers illustrated in Fig. 8. All blocks featured a square base with side length L = 50 mm. Table 1 provides detailed parameters for each block, where *Offset* is the distance from each block's contact surface center to the tray center; and $\mu$ is the friction coefficient.

**Table 1** Height, offset distance, friction coefficient and numbering of experimental objects

| Offset(mm) | $\mu$ | Height(mm) | | |
|---|---|---|---|---|
| | | 50 | 150 | 250 |
| 100 | 0.15 | a1 | b1 | c1 |
| | 0.3 | A1 | B1 | C1 |
| 200 | 0.15 | a2 | b2 | c2 |
| | 0.3 | A2 | B2 | C2 |
| 300 | 0.15 | a3 | b3 | c3 |
| | 0.3 | A3 | B3 | C3 |

Each experimental object had a mass of 1 kg, with mass concentrated at the 8 vertices to achieve maximum rotational inertia. The inertia matrix was expressed as:

$$I_{obj} = \text{diag}\left(0.25\left(L^2 + \text{Height}^2\right),\quad 0.25\left(L^2 + \text{Height}^2\right),\quad 0.5L^2\right) \tag{16}$$

The performance of the trajectory smoother was evaluated using a 3D cosine trajectory corrupted with white noise as the test input. To accommodate the operational workspace of the manipulator, the trajectory was parameterized as follows:

$$\begin{cases} x_m(t) = \tilde{x}_m(t) + n(t),\ 0 \leqslant t \leqslant \text{T}, \\ \tilde{x}_m(t) = \begin{bmatrix} \tilde{x}_{m,1}(t) & \tilde{x}_{m,2}(t) & \tilde{x}_{m,3}(t) \end{bmatrix}^{\text{T}}, n(t) = \begin{bmatrix} n_1(t) & n_2(t) & n_3(t) \end{bmatrix}^{\text{T}}, \\ \tilde{x}_{m,1}(t) = 0.3\left(1 - \cos(2\pi/5 \cdot t)\right),\ \tilde{x}_{m,2}(t) = 0.2\left(1 - \cos(2\pi/6 \cdot t)\right), \\ \tilde{x}_{m,3}(t) = 0.1\left(1 - \cos(2\pi/7 \cdot t)\right), \\ n_1(t) \sim N\left(0, 0.01^2\right),\ n_2(t) \sim N\left(0, 0.007^2\right),\ n_3(t) \sim N\left(0, 0.0003^2\right). \end{cases}$$

The simulation employed a control frequency of 50 Hz with a total simulation time of $\text{T} = 600\text{s}$. This duration ensured that differences in initial object positions did not affect the results when comparing objects with identical offset distances.

The experimental evaluation employed two control paradigms:

F-mode (Filtering-only): The user input trajectory $x_m(t)$ is processed through filtering to produce reference position $x_r(t)$, and let $x_d(t) = x_r(t)$. The reference acceleration $\ddot{x}_d(t)$ results from second-order differentiation of $x_d(t)$, from which the desired manipulator orientation is determined. Manipulator control employs inverse kinematics as specified in Section 3.

FSC-mode (Filtering and Smoothing with Constraints): The proposed approach featuring the control flow depicted in Fig. 6. This mode is equivalent to MPC-based trajectory smoothing subject to dynamic constraints, carried out on the basis of the F mode.

Both control modes utilized a reference object O positioned at the tray center and configured with identical geometry to object c3. Given that all experimental objects employed maximum inertia matrix values, the reference object inertia matrix matched that of c3. The input trajectory $x_m$ and output trajectory $x_s$ are the desired and actual trajectories of reference object O, respectively. Both modes employed an identical filter with transfer function $H(z) = \dfrac{1\mathrm{e}^{-3}\left(3.6 + 7.2z^{-1} + 3.6z^{-2}\right)}{1 - 1.8228z^{-1} + 0.8372z^{-2}}$. In FSC-mode, the distance between offset object A and reference object O was $\|P_A\| = 325\,\text{mm}$, with the friction coefficient between offset object and tray set to $\mu = 0.15$. The MPC prediction horizon was N = 10.

Consider the following three metrics:

(1) The FC stability metric $S^{obj}(t)$, which represents the ratio of the contact force angle to the friction angle, $S^{obj}(t) = \arccos\left(|F_{c,3}^{obj}|/\|F_c^{obj}\|\right)/\arctan(\mu)$, where $F_c^{obj} \in \mathbb{R}^3$ is the contact force vector applied to experimental object *obj* in its CoM coordinate frame, $F_c^{obj} = \begin{bmatrix} F_{c,1}^{obj} & F_{c,2}^{obj} & F_{c,3}^{obj} \end{bmatrix}^\mathrm{T}$. To ensure smooth conduct of experiments in both control modes, each experimental object is rigidly connected to the tray via force sensors in the simulation, with $F_c^{obj}$ measured through these sensors. The stability criteria are: $S^{obj}(t) \leqslant 1$ indicates the contact force lies within the FC (no sliding), while $S^{obj}(t) > 1$ signifies FC violation (sliding occurs). $S_{\max}^{obj}$ is the maximum value of $S^{obj}(t)$ observed during the experiment, $S_{\max}^{obj} = \max\{S^{obj}(t), 0 \leqslant t \leqslant \mathrm{T}\}$. For different experimental objects, a smaller $S_{\max}^{obj}$ indicates reduced sliding tendency.

(2) The stability metric $B^{obj}(t)$, which quantifies the relationship between the ZMP position and the contact surface boundaries, $B^{obj}(t) = \dfrac{2}{d}\sqrt{\left(P_{zero,1}^{obj}\right)^2 + \left(P_{zero,2}^{obj}\right)^2}$, where $P_{zero}^{obj}$ is the ZMP of the experimental object in its CoM coordinate frame, $P_{zero}^{obj} = \begin{bmatrix} P_{zero,1}^{obj} & P_{zero,2}^{obj} & P_{zero,3}^{obj} \end{bmatrix}^\mathrm{T}$; $d$ is the diameter of the contact surface between the object and the tray, $d = \mathrm{L}$. The stability criteria are: $B^{obj}(t) \leqslant 1$ indicates the ZMP lies within the contact surface boundary (stable equilibrium), while $B^{obj}(t) > 1$ signifies the ZMP falls outside the contact surface (potential tipping). For computational purposes, $B^{obj}(t) > 1$ is treated as an unstable condition. The maximum stability parameter $B_{\max}^{obj}$ is the peak instability during the experiment, $B_{\max}^{obj} = \max\{B^{obj}(t), 0 \leqslant t \leqslant \mathrm{T}\}$. Among different experimental objects, a smaller $B_{\max}^{obj}$ indicates enhanced stability against tipping.

(3) The trajectory tracking error, which quantifies the deviation of the actual trajectory from the noise-free cosine reference trajectory: $E(t) = \|x_s(t) - \tilde{x}_m(t)\|$. The average trajectory error is:

$$\bar{E} = \frac{1}{\mathrm{T}} \int_0^\mathrm{T} E(t)\,\mathrm{d}t.$$

### 4.2 Simulation Results

The $S_{\max}^{obj}$ and $B_{\max}^{obj}$ values for experimental objects under both control modes are presented in Fig. 9. The F-mode results far exceed the safety threshold, whereas the FSC-mode successfully maintains $S_{\max}^{obj}$ and $B_{\max}^{obj}$ below unity for all objects while preserving adequate safety margins. This validates the effectiveness of the proposed method in enforcing object constraint satisfaction.

Several key relationships are revealed in Fig. 9 (b) and (d):

• Offset distance effect: For objects at the same height, larger offset distances result in higher $S_{max}^{obj}$ and $B_{max}^{obj}$ values.

• Height effect: For identical offset distances, shorter objects exhibit both smaller rotational inertia and correspondingly lower $B_{max}^{obj}$ values, while simultaneously creating larger CoM distances from the reference object and resulting in higher $S_{max}^{obj}$ values.

• Friction coefficient effect: For objects with identical geometry, reduced friction coefficients yield higher $S_{max}^{obj}$ values.

These observations align with the theoretical analysis presented in Section 2.3, confirming the multi-object dynamic constraint framework.

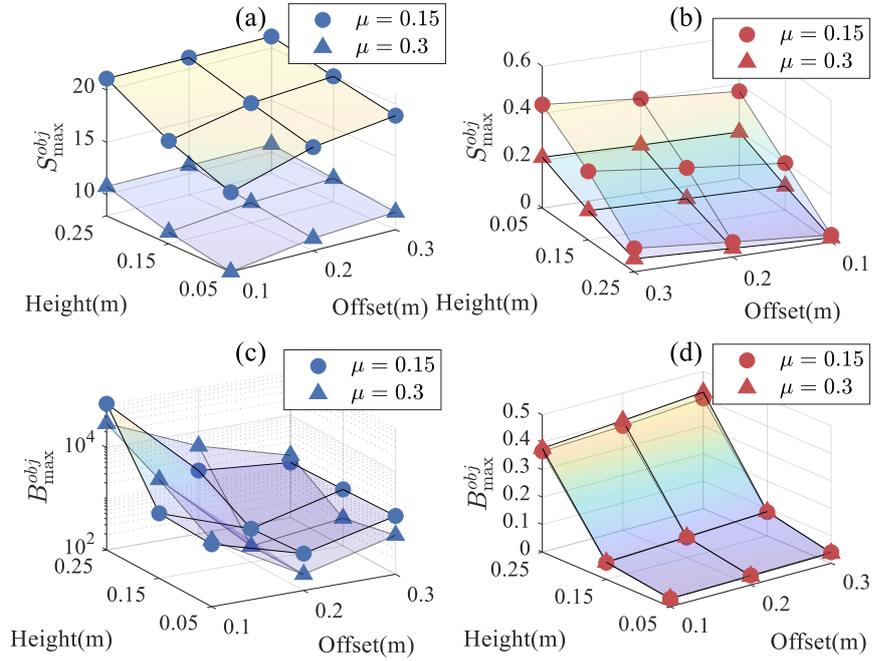

**Fig. 9** Simulation results. (a) and (c) represent experimental object $S_{max}^{obj}$ and $B_{max}^{obj}$ under F-mode, respectively. (b) and (d) represent experimental object $S_{max}^{obj}$ and $B_{max}^{obj}$ under FSC-mode, respectively.

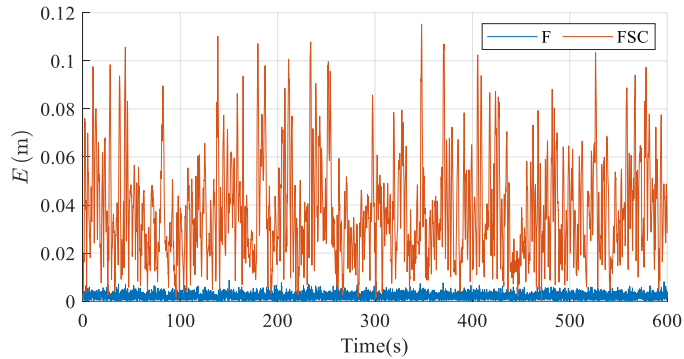

**Fig. 10** Trajectory tracking errors for both control modes.

Although the FSC-mode effectively guarantees object dynamic constraint satisfaction, it compromises trajectory tracking accuracy. As shown in Fig. 10, the FSC-mode exhibits a maximum trajectory error of 0.115 m and an average trajectory error of 0.0347 m, whereas the F-mode achieves a maximum trajectory error of 0.008 m and an average trajectory error of 0.0022 m. This performance trade-off is expected, as the F-mode tracking errors originate solely from filtering and inverse kinematics operations, while the FSC-mode introduces additional trajectory smoothing and constraint enforcement. In the designed cascaded five-integrator position control system, the FSC constraints effectively limit the system's driving capability, which necessarily results in larger trajectory errors.

While high trajectory tracking accuracy is beneficial, maintaining object dynamic constraints remains the primary objective during nonprehensile transportation tasks. The experiments employed identical filters for both control modes to evaluate the trajectory smoothing effects on performance outcomes. Although filters objectively smooth input trajectories, experimental comparisons demonstrate that filtering alone cannot provide definitive constraint guarantees. While the F-mode could achieve improved performance through lower cutoff frequency filtering of input trajectories, such approaches cannot theoretically satisfy state constraints for nonprehensile transportation tasks, thus failing to ensure reliable task completion. In contrast, the FSC-mode incorporates filters solely for differential computation of input trajectory velocities and accelerations required for feedback control. When master devices directly provide user input velocities and accelerations, the filter becomes unnecessary.

## 5 Experiments and Results

Experiments were conducted to evaluate the actual performance of the algorithm by controlling a physical robot through the teleoperation framework. The teleoperation system acquired the user's 3D positional input via the master device. The user involvement was restricted to the positional control of the tray, while the proposed algorithm autonomously controlled the tray's orientation and ensured that objects on the tray neither slide nor tip over during manipulation.

### 5.1 Experimental Setup

The experimental scenario is shown in Fig. 11. The teleoperation system employed a CyberSystem hand controller as the master device to track positional data from the user's wrist and incrementally map the user's wrist position to the robotic arm's workspace within Cartesian space. The slave system consisted of a Franka Emika Panda 7-degree-of-freedom (DoF) robotic arm. To accommodate the extensive workspace of the robotic arm, user displacement inputs were scaled by a factor of 2 before being mapped to the robot's workspace. The robot was operated in the joint impedance control mode, with the stiffness of all joints set to the maximum allowable value of 14000 Nm/rad. An aluminum plate serving as a tray was mounted at the robot arm's end effector. Multiple objects to be transported were randomly placed on the tray, which had a radius of $R_{tray} = 200$ mm. The experimental objects comprised common items such as vases, water bottles, wooden blocks, and glassware. Object heights ranged from 100mm to 235mm, with their shapes, dimensions, and numbers shown in Fig. 12. Since the inevitable sliding displacement of objects during manipulation needed to be monitored, markers were affixed to the top of each object to enable camera-based detection. A camera system mounted above the tray via a rigid bracket captured the sliding displacement of the objects throughout the experimental trials. It should be noted that the camera is solely employed to measure the sliding distance of objects for result analysis, whereas the control of object positions is implemented in an open-loop manner.

A computer running Ubuntu 20.04 with Intel Core i9-9900K CPU and 64GB RAM was used in the experiments to run the teleoperation controller. The controller was implemented in C++ and employed multithreading to separate the processes of master device data acquisition, teleoperation control, and robot motion control. The data acquisition thread collected user input from the master device at 250 Hz, with a communication latency of less than 1 ms between the master device and the teleoperation controller. The teleoperation control ran at 50 Hz in the planning thread, outputting interpolated desired motions to a data queue. The optimization problem was solved using the open-source qpOASES package, with an average solution time of 7.6 ms per control cycle (standard deviation ±1.8 ms). The hardware interface thread handled communication with the robotic arm. In each control cycle, it retrieved the robot state feedback, read the desired motion commands from the data queue, and controlled the robot accordingly. The communication frequency between this thread and the robot was 1 kHz, with a latency below 1 ms.

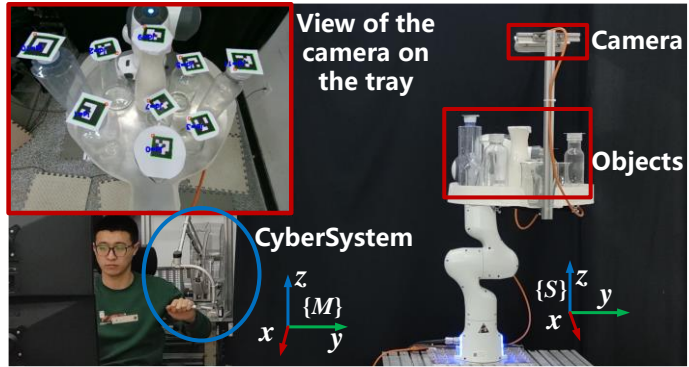

**Fig. 11** Schematic diagram of the experimental scenario. The master device operated by the user is a CyberSystem hand controller, and the user's displacement is scaled by a factor of 2 and mapped to the slave robot's workspace. Objects on the tray have target markers affixed to their tops. The top left corner shows the camera's field of view mounted on the tray, which detects the sliding displacement of the objects.

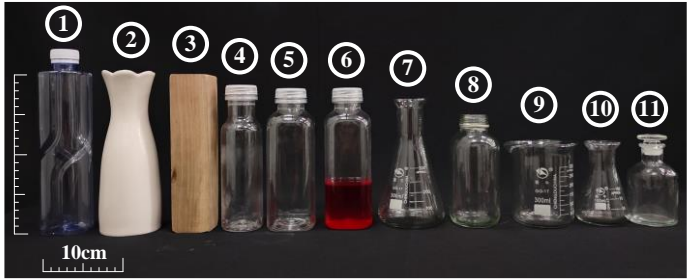

**Fig. 12** Objects used in the experiment and their numbers.

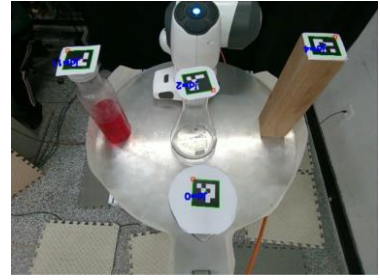

**Fig. 13** Position of objects in experiments F4 and FSC4

Targeted experiments were conducted to validate the robustness of the proposed method against uncertainties in object parameters, as shown in Table 2. These experiments varied the quantity, shape, position, and friction coefficient of the objects. The rationale for this focus on robustness stems from the algorithm's operational mechanics: it relies on estimating object shape parameters to inform the trajectory smoother's constraints. Acknowledging that such estimations are inherently prone to error, the method was intentionally developed to function effectively even in the presence of these inaccuracies. Therefore, the experiments were essential to confirm that the proposed approach is not rigidly dependent on precise parameter estimation and maintains its performance under more realistic conditions.

**Table 2** Detailed experimental design

| Exp. No. | Object No. | $H$ | $\|P_A\|$ | $\mu$ |
|---|---|---|---|---|
| F4 | 3, 6, 7, 9 | 200mm | -- | -- |
| FSC4 | 3, 6, 7, 9 | 200mm | 224mm | 0.2 |
| F9 | Except for 3,6 | 200mm | -- | -- |
| FSC9 | Except for 3,6 | 180mm | 220mm | 0.3 |

Two control modes were implemented: F-mode and FSC-mode. In the experiment nomenclature, letters indicate the control mode while numbers denote the quantity of objects manipulated.

Experiments F4 and FSC4 utilized identical object positions as shown in Fig. 13, while experiments F9 and FSC9 utilized identical object positions as illustrated in the upper left corner of Fig. 11.

The transfer functions for the two control modes were defined as follows: $H_F(z) = \dfrac{1e^{-2}(2.01 + 4.02z^{-1} + 2.01z^{-2})}{1 - 1.561z^{-1} + 0.6414z^{-2}}$ for F-mode and $H_{FSC}(z) = \dfrac{0.0976 + 0.1953z^{-1} + 0.0976z^{-2}}{1 - 0.9428z^{-1} + 0.3333z^{-2}}$ for FSC-mode. Both control modes established a reference object O as a rectangular block with a square base of 50mm side length and height $H$. The reference object was positioned at the center of the tray. In FSC-mode, a offset object A was defined with distance $\|P_A\| = \sqrt{R_{tray}^2 + (0.5H)^2}$ from the reference object O, whose inertia matrix was calculated according to Eq. (16). The friction coefficient between the offset object and the tray was denoted as $\mu$. For experiment FSC4, all four objects had friction coefficients exceeding 0.2, with Object No. 3 being a homogeneous rectangular block most susceptible to tipping. The offset object parameters were configured to match the characteristics of Object No. 3, ensuring accurate parameter estimation.

Experiment FSC9 simulated scenarios with inaccurate offset object parameter estimation. Objects No. 1, 2, 4, and 5 had heights exceeding 180mm, with Object No. 1 being the tallest at 235mm. Objects No. 7, 8, and 10 had friction coefficients below 0.3, with Object No. 10 having the lowest friction coefficient of 0.21. All other FSC-mode parameters remained consistent with simulation settings.

To effectively evaluate experimental performance, all four experimental groups used the same input trajectories. First, the teleoperation controller employed the control parameters from experiment FSC4. The operator provided 3D positional inputs via the CyberSystem hand controller, controlling only the translational position of the tray, while the teleoperation controller autonomously regulated the tray's orientation to ensure the stability of the objects. Throughout the experiment, the operator's input trajectories were recorded in real time. Subsequently, the recorded input trajectories were reused for the remaining three experimental groups, with controller parameters adjusted accordingly.

To minimize experimental variability, a total of 10 participants were recruited. All participants signed informed consent forms prior to the experiment. During the experiment, only the operator's input trajectories were recorded; no sensitive or privacy-related data were collected. The study was approved by the institutional ethics committee and conducted in accordance with the Declaration of Helsinki. Among the participants, one was female and nine were male, with one participant having prior teleoperation experience. All participants demonstrated normal device operation capabilities. Before commencing the experiment, the experimenters provided each participant with 15 minutes of instruction on the experimental tasks and assisted them in familiarizing themselves with the devices. Each participant completed two trials, producing two recorded input trajectories. All input trajectories exhibited maximum velocities greater than 0.8 m/s and maximum accelerations exceeding 2.4 m/s² (after mapping to the robot workspace), covering the entire robot workspace. Each trajectory lasted approximately 50 seconds. In total, 20 distinct user input trajectories were recorded, resulting in 80 experimental trials. The following performance metrics were evaluated:

(1) Trajectory tracking error $E(t) = \|x_s(t) - x_m(t)\|$, representing robot actual trajectory error from input trajectories. The average trajectory tracking error is calculated as: $\bar{E} = \dfrac{1}{T}\int_0^T E(t)\,dt$, where T is the trajectory duration.

(2) Object sliding distance $D^{obj}(t)$, obtained by detecting the position of the target by the camera. $D^{obj}_{\max}$ is the maximum sliding distance of the object.

**5.2 Experimental Results**

The average trajectory tracking error distributions across all experiments are presented in Table 3. Experiments F4 and F9 have employed identical control parameters, generating identical output trajectories for corresponding input trajectories. In F-mode operation, a delay of 0.12 s is observed in the desired trajectory relative to the input trajectory, attributed to filter processing. When delay effects are excluded, F-mode exhibits average trajectory error is 4.5 mm with highly concentrated error distributions. These trajectory errors primarily originate from filter processing and inverse kinematics calculations. FSC-mode has the same total delay as F-mode (0.12 s), comprising 0.03 s from filter processing and 0.09 s from trajectory smoothing. The reduced filter delay results from the higher cutoff frequency employed in FSC-mode compared to F-mode. Excluding delay effects, experiments FSC4 and FSC9 achieve average trajectory errors of 15.5 mm and 12.5 mm, respectively. The discrepancy between these experimental results is attributed to the dynamic constraints.

**Table 3** The mean and standard deviation of $\bar{E}$.

| Exp. No. | Mean(mm) | Standard deviation(mm) |
|---|---|---|
| F4&F9 | 4.5 | 0.17 |
| FSC4 | 15.5 | 1.93 |
| FSC9 | 12.5 | 1.91 |

The dynamic constraint profiles for FSC4 and FSC9 during the 10-20 second interval for a representative input trajectory are illustrated in Fig. 14 (c). FSC4 imposes tighter constraints due to more conservative offset object parameter settings, which correspondingly limits the system's dynamic response capability and results in relatively larger trajectory errors. Nevertheless, the tracking accuracy achieved by the FSC-mode remains sufficient for effective task execution.

The sliding distance distributions for Objects No. 3, 6, 7, and 9 in experiments F4 and FSC4 are presented in Fig. 14 (a). The mean sliding distances are 9.40 mm for F4 and 2.59 mm for FSC4, representing a notable reduction of 72.45% with FSC-mode control ($p < 0.001$). Across all input trajectories, individual object sliding distances in experiment FSC4 consistently remain shorter than those in experiment F4. The four selected objects feature diverse characteristics: varying friction coefficients, predominantly irregular geometries (except for Object No. 3), and different physical properties, with Object No. 6 being a water bottle partially filled with water. One-way ANOVA analysis ($\alpha = 0.05$) reveals no significant differences in sliding distances among the four objects ($F(3,76) = 0.51$, $p = 0.677$). Notably, no objects tipped over throughout any experimental trials. These results demonstrate that the proposed algorithm effectively maintains dynamic constraints while exhibiting adaptability to variations in object geometry, positions, and friction coefficients.

Although objects in experiment F4 also did not tip over, they produced larger sliding distances. To elucidate this phenomenon, Fig. 14 (d) and (e) present the output orientation trajectories for both control modes during the 10-20 second interval using the same input trajectory as shown in Fig. 14 (c). The orientation trajectories generated by FSC-mode show superior smoothness compared to those of F-mode. This enhanced smoothness results from the MPC-based trajectory smoother, which constrains the fourth-order derivatives of position trajectories, thereby preventing abrupt changes in output orientation and reducing the likelihood of object sliding and oscillation. The maximum tilt angle of objects during both experiments exceeded 13.7°. The supplementary video (https://youtu.be/NsWn40RT-I0) reveals that liquid surface oscillations in Object No. 6 and camera field-of-view vibrations were more pronounced in experiment F4 compared to FSC4, further confirming the inferior trajectory smoothness characteristic of F-mode operation.

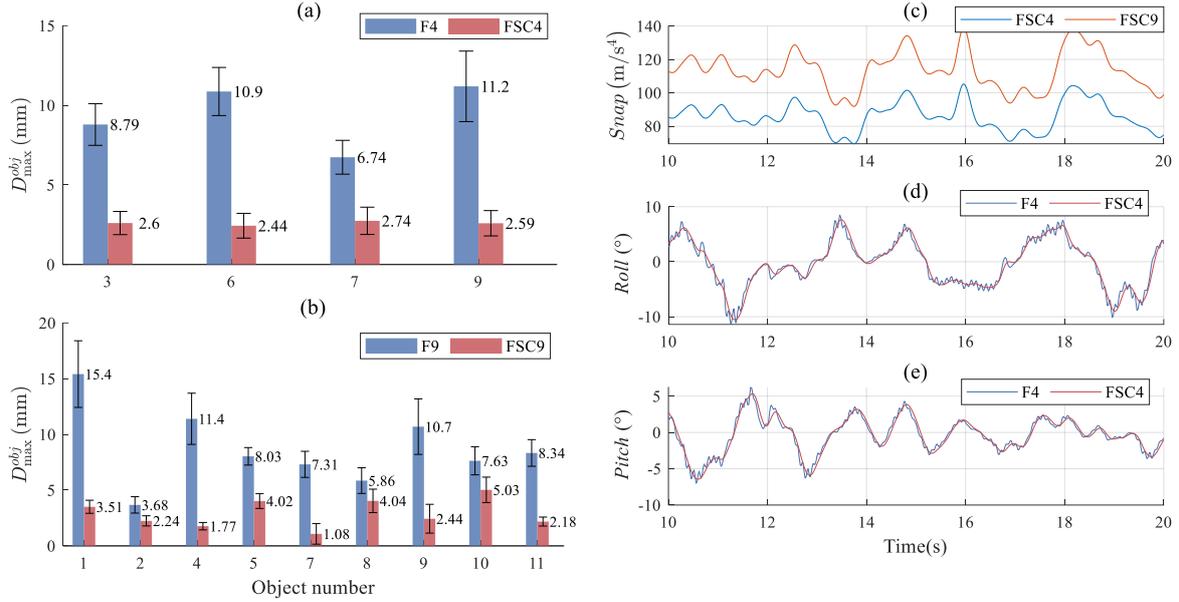

**Fig. 14** Experimental results for four control configurations. (a) Sliding distance distributions for individual objects in experiments F4 and FSC4. (b) Sliding distance distributions for individual objects in experiments F9 and FSC9. (c) Dynamic constraint profiles for experiments FSC4 and FSC9 during 10-20 s interval using a representative input. (d, e) Output orientation trajectories for experiments F4 and FSC4 during 10-20 s interval using the same input trajectory as (c).

In experiment FSC9, no objects tipped over across all input trajectories. Conversely, experiment F9 exhibited a 13.9% object tip-over rate, with the following distribution: Object No. 1 tipped over 13 times, Object No. 4 tipped over 10 times, Object No. 5 tipped over 2 times, while other objects remained stable. Additionally, Object No. 11 was occluded by tipped-over objects on 2 occasions. After excluding all tip-over and occlusion events from the analysis, the sliding distance distributions for individual objects are presented in Fig. 14 (b). Objects in experiment FSC9 demonstrate substantially shorter sliding distances compared to those in F9 ($p < 0.001$). In FSC9, the maximum sliding distance is 7.47 mm, with 93.3% of objects moving less than 5 mm. In contrast, F9 shows a maximum sliding distance of 20.78 mm, with 85.1% of objects exceeding 5 mm sliding distance.

It should be noted that the FSC-mode requires approximately 7.6 ms to solve the optimization problem at each control step, whereas the computational cost of the F-mode is almost negligible. Although the FSC-mode introduces additional computational overhead, it guarantees dynamic constraint satisfaction and $C^4$-continuous trajectories, which cannot be achieved by simple filtering alone. Therefore, its performance benefits clearly outweigh the associated computational cost.

In addition, some of the experimental objects were containers filled with liquid. The theoretical model assumes rigid bodies with fixed centers of mass and constant dynamics, whereas liquid sloshing alters the CoM distribution and introduces additional dynamics. We deliberately chose liquid containers for the experiments because, on one hand, the surface oscillations provide a visual indicator of trajectory smoothness, highlighting differences in object motion stability between algorithms; on the other hand, the sloshing introduces natural variations in overall object dynamics, offering a realistic test of the algorithm's robustness to inertial disturbances. Since the amplitude of liquid motion is limited, the internal fluid dynamics were not explicitly modeled. Experimental results show that, despite liquid sloshing, no objects tipped over, and sliding distances remained within acceptable limits, demonstrating the algorithm's robustness to inertial perturbations.

The progression from experiment FSC4 to FSC9 represents a transition toward real-world applications, incorporating increased complexity through several factors: expanded object quantities, enhanced shape diversity, varied friction coefficients, and dispersed object distributions across the entire tray surface. These conditions impose higher requirements on algorithm applicability. Experiment FSC9 simulates scenarios where parameter estimation for offset objects is inaccurate. The results confirm that the proposed method remains applicable to multi-object nonprehensile transportation tasks under conditions of 23.4% shape deviation and 30% friction coefficient deviation for the offset object. Across all input trajectories, objects do not tip over and show shorter sliding distances compared to experiment F9, highlighting the method's robustness to shape and friction coefficient variations.

## 6 Discussion

Both simulations and experiments involve objects with friction coefficients greater than 0.15 and height-to-width ratios below 6. The proposed method does not prove applicable for objects with very low friction coefficients ($\mu < 0.1$) or large height-to-width ratios ($H/d > 8$). Such conditions impose more stringent dynamic constraints and significantly increase trajectory tracking errors. From a practical perspective, achieving friction coefficients below 0.1 under dry friction conditions with direct contact is difficult for common materials [52]. Self-lubricating materials specifically designed for friction reduction are inherently unsuitable for nonprehensile transportation modes. Furthermore, for objects with height-to-width ratios exceeding 6, even when transport is required, they are not deliberately positioned upright without stabilization measures. Therefore, the proposed method is sufficient for common operational conditions.

Additional experiments were conducted using online teleoperation rather than pre-recorded trajectories. Users provided movement commands to the robot in real-time. The supplemental video demonstrates the experimental procedure over approximately 3 minutes of operation, with maximum acceleration reaching 2.2 m/s² and tray tilt angles exceeding 12.5°. Results show that objects experienced a maximum sliding distance of 10.3 mm, with no objects tipping over throughout the experiment.

### 6.1 Limitations and Future Work

First, this research primarily focuses on scenarios where multiple objects are distributed on a planar surface, without considering stacked configurations. In stacked arrangements, inter-object contact dynamics become considerably more complex, representing a critical challenge for future research. Second, given the emphasis on task-space constraints and control, the present algorithm does not yet rigorously integrate low-level joint limits, velocity, and acceleration constraints of the robot into the solver. Nevertheless, experimental results indicate that the system can operate stably within the robot's physical performance envelope for conventional transportation tasks. Decoupling low-level joint constraints and incorporating them into the control framework represents a promising direction for further enhancing system performance under complex constraint conditions.

Future work will focus on extending this method to the complex scenarios of manipulating stacked objects and exploring the integration of stricter joint-space constraints into the control framework, thereby further perfecting this shared teleoperation system.

## 7 Conclusions

This work addresses teleoperation challenges in multi-object nonprehensile transportation by proposing a VO-MPC-based shared teleoperation method. The approach innovatively introduces virtual reference objects and virtual offset objects to transform multi-object dynamic constraints into FC and ZMP constraints for a single offset object, significantly reducing model dependence. Based on this constraint transformation, a dynamic constraint criterion for trajectory planning is established. The criterion is integrated with an MPC-based trajectory smoother that generates $C^4$ continuous trajectories while coordinating target tracking and tray orientation control.

Simulation results indicate that the proposed FSC-mode ensures all objects satisfy constraints (both FC and ZMP metrics remain below 1) with adequate safety margins, while the baseline F-mode significantly violates these constraints. The relationship between these metrics and object properties validates the effectiveness of the multi-object dynamic constraint processing approach. Experimental validation further confirms these findings: the FSC-mode successfully handles 9 different items at accelerations up to 2.4 m/s², achieving 72.45% reduction in sliding distance and decreasing tip-over rate from 13.9% to 0% compared to F-mode. The system maintains safe operation even with 23.4% height deviation and 30% friction coefficient deviation, demonstrating robustness across variations in object parameters.

In summary, this study provides a practical processing method for multi-object dynamic constraints in nonprehensile manipulation tasks and offers a viable solution for implementing teleoperation control. The proposed method exhibits strong applicability and robustness, showing potential for improving manipulation efficiency and reducing the risk of item damage.

**Appendixes**

**Appendix A.** Relationship between Angular Acceleration Constraints and the Fourth-Order Derivative of Position

Referring to Fig. 2, the orientation of object A is expressed as $\Phi_d = \alpha \hat{n}$, as described in Eq. (1). The angular velocity of object A is given by:

$$\Omega_A = \frac{d\Phi_d}{dt} = \frac{d}{dt}(\alpha \hat{n}) = \dot{\alpha}\hat{n} + \alpha \frac{d\hat{n}}{dt}. \tag{17}$$

The angular acceleration of object A is:

$$\dot{\Omega}_A = \frac{d^2\Phi_d}{dt^2} = \frac{d^2}{dt^2}(\alpha \hat{n}) = \frac{d}{dt}\left(\dot{\alpha}\hat{n} + \alpha \frac{d\hat{n}}{dt}\right) = \ddot{\alpha}\hat{n} + 2\dot{\alpha}\frac{d\hat{n}}{dt} + \alpha \frac{d^2\hat{n}}{dt^2}. \tag{18}$$

The terms $\alpha \frac{d\hat{n}}{dt}$ in Eq. (17) and $2\dot{\alpha}\frac{d\hat{n}}{dt} + \alpha \frac{d^2\hat{n}}{dt^2}$ in Eq. (18) account for changes in the rotation axis. Where,

$$\frac{d\hat{n}}{dt} = \frac{\dddot{x}_d \times g}{\|\ddot{x}_d \times g\|} - \frac{(\ddot{x}_d \times g)^T (\dddot{x}_d \times g)(\ddot{x}_d \times g)}{\|\ddot{x}_d \times g\|^3}, \tag{19}$$

$$\frac{d^2\hat{n}}{dt^2} = \frac{\ddddot{x}_d \times g}{\|\ddot{x}_d \times g\|} - \frac{2(\ddot{x}_d \times g)^T (\dddot{x}_d \times g)(\dddot{x}_d \times g)}{\|\ddot{x}_d \times g\|^3} + \frac{3\left((\ddot{x}_d \times g)^T (\dddot{x}_d \times g)\right)^2 (\ddot{x}_d \times g)}{\|\ddot{x}_d \times g\|^5}$$
$$- \frac{(\dddot{x}_d \times g)^T (\dddot{x}_d \times g)(\ddot{x}_d \times g)}{\|\ddot{x}_d \times g\|^3} - \frac{(\ddot{x}_d \times g)^T (\ddddot{x}_d \times g)(\ddot{x}_d \times g)}{\|\ddot{x}_d \times g\|^3}. \tag{20}$$

Substituting Eqs. (1), (17), (19), and (20) into Eq. (18) and simplifying yields:

$$\dot{\Omega}_A = \frac{\ddddot{x}_d \times g}{\|\ddot{x}_d \times g\|}\alpha + 2\frac{\dddot{x}_d \times g}{\|\ddot{x}_d \times g\|}\dot{\alpha} - 2\frac{(\ddot{x}_d \times g)^T (\dddot{x}_d \times g)(\ddot{x}_d \times g)}{\|\ddot{x}_d \times g\|^3}\alpha + \frac{\ddot{x}_d \times g}{\|\ddot{x}_d \times g\|}\ddot{\alpha}$$
$$- 2\frac{(\ddot{x}_d \times g)^T (\dddot{x}_d \times g)(\ddot{x}_d \times g)}{\|\ddot{x}_d \times g\|^3}\dot{\alpha} - \frac{(\dddot{x}_d \times g)^T (\dddot{x}_d \times g)(\ddot{x}_d \times g)}{\|\ddot{x}_d \times g\|^3}\alpha$$
$$- \frac{(\ddot{x}_d \times g)^T (\ddddot{x}_d \times g)(\ddot{x}_d \times g)}{\|\ddot{x}_d \times g\|^3}\alpha + 3\frac{\left((\ddot{x}_d \times g)^T (\dddot{x}_d \times g)\right)^2 (\ddot{x}_d \times g)}{\|\ddot{x}_d \times g\|^5}\alpha.$$

Denoting the terms excluding the first as:

$$\dot{\Omega}_{A,1} = 2\frac{\dddot{x}_d \times g}{\|\ddot{x}_d \times g\|}\dot{\alpha} - 2\frac{(\ddot{x}_d \times g)^{\mathrm{T}}(\dddot{x}_d \times g)(\ddot{x}_d \times g)}{\|\ddot{x}_d \times g\|^3}\alpha + \frac{\ddot{x}_d \times g}{\|\ddot{x}_d \times g\|}\ddot{\alpha}$$

$$-2\frac{(\ddot{x}_d \times g)^{\mathrm{T}}(\ddot{x}_d \times g)(\ddot{x}_d \times g)}{\|\ddot{x}_d \times g\|^3}\dot{\alpha} - \frac{(\ddot{x}_d \times g)^{\mathrm{T}}(\dddot{x}_d \times g)(\ddot{x}_d \times g)}{\|\ddot{x}_d \times g\|^3}\alpha$$

$$-\frac{(\ddot{x}_d \times g)^{\mathrm{T}}(\ddot{x}_d \times g)(\dddot{x}_d \times g)}{\|\ddot{x}_d \times g\|^3}\alpha + 3\frac{\left((\ddot{x}_d \times g)^{\mathrm{T}}(\ddot{x}_d \times g)\right)^2(\ddot{x}_d \times g)}{\|\ddot{x}_d \times g\|^5}\alpha.$$

Experimental observations indicate that, under the maximum tray acceleration (2.4 m/s²) and maximum tilt angle (13.7°) considered in this study, the contribution of $\dot{\Omega}_{A,1}$ is negligible ($\|\dot{\Omega}_{A,1}\| < 0.05\|\dot{\Omega}_A\|$). Therefore, $\dot{\Omega}_A$ can be approximated as:

$$\dot{\Omega}_A \approx \frac{\dddot{x}_d \times g}{\|\ddot{x}_d \times g\|}\alpha.$$

Further, considering the case when $\alpha = 0$.

$$\alpha = \arccos\left(\frac{g^{\mathrm{T}}(g - \ddot{x}_d)}{\|g\|\|g - \ddot{x}_d\|}\right) = \arccos\left(\frac{g_3(g_3 - \ddot{x}_{d,3})}{|g_3|\sqrt{\ddot{x}_{d,1}^2 + \ddot{x}_{d,2}^2 + (g_3 - \ddot{x}_{d,3})^2}}\right),$$

$$\|\ddot{x}_d \times g\| = \left|g_3\sqrt{\ddot{x}_{d,1}^2 + \ddot{x}_{d,2}^2}\right|,$$

where $\ddot{x}_d = [\ddot{x}_{d,1}\ \ \ddot{x}_{d,2}\ \ \ddot{x}_{d,3}]^{\mathrm{T}}$, $g = [0\ \ 0\ \ g_3]^{\mathrm{T}}$. When the object's horizontal-plane acceleration is zero ($\sqrt{\ddot{x}_{d,1}^2 + \ddot{x}_{d,2}^2} = 0$), $\alpha = 0$ and $\|\ddot{x}_d \times g\| = 0$. Let $\ddot{x}_{d,12} = \sqrt{\ddot{x}_{d,1}^2 + \ddot{x}_{d,2}^2}$, then we have:

$$\lim_{\ddot{x}_{d,12} \to 0}\frac{\alpha}{\|\ddot{x}_d \times g\|} = \lim_{\ddot{x}_{d,12} \to 0}\frac{d\alpha}{d\ddot{x}_{d,12}}\bigg/\frac{d\|\ddot{x}_d \times g\|}{d\ddot{x}_{d,12}} = \lim_{\ddot{x}_{d,12} \to 0}\frac{g_3(g_3 - \ddot{x}_{d,3})}{|g_3|^2\left(\ddot{x}_{d,12}^2 + (g_3 - \ddot{x}_{d,3})^2\right)} = \frac{1}{\|g\|\|\ddot{x}_d - g\|}.$$

Thus, the relationship between the fourth-order derivative of position and the angular acceleration of the object is expressed as:

$$\dot{\Omega}_A \approx \begin{cases} \dfrac{\dddot{x}_d \times g}{\|\ddot{x}_d \times g\|}\alpha = \dfrac{\alpha}{\|\ddot{x}_d \times g\|}\begin{bmatrix}g_3\dddot{x}_{d,2} & -g_3\dddot{x}_{d,1} & 0\end{bmatrix}^{\mathrm{T}}, \alpha \neq 0, \\ \dfrac{\dddot{x}_d \times g}{\|g\|\|\ddot{x}_d - g\|} = \dfrac{1}{\|g\|\|\ddot{x}_d - g\|}\begin{bmatrix}g_3\dddot{x}_{d,2} & -g_3\dddot{x}_{d,1} & 0\end{bmatrix}^{\mathrm{T}}, \alpha = 0. \end{cases}$$

Consequently, the constraint on object angular acceleration can be transformed into a constraint on the fourth-order derivative of the object's position:

$$\|\dddot{x}_d\| \leqslant \begin{cases} \|\dot{\Omega}_A\|_{\max}\dfrac{\sqrt{\ddot{x}_{d,1}^2 + \ddot{x}_{d,2}^2}}{\alpha}, \alpha \neq 0, \\ \|\dot{\Omega}_A\|_{\max}|g_3 - \ddot{x}_{d,3}|, \alpha = 0. \end{cases}$$

**Nomenclature**

Abbreviations

| | |
|---|---|
| VO | Virtual object |
| MPC | Model predictive control |
| FC | Friction cone |
| ZMP | Zero-Moment Point |
| IL | Imitation Learning |

| | | |
|---|---|---|
| RL | Reinforcement Learning | |
| QP | Quadratic programming | |
| 3D | Three-dimensional | |
| CoM | Center of mass | |
| DoF | Degree of freedom | |

Variables

| | | |
|---|---|---|
| $F_S$ | Supporting force of the object |
| $F_g$ | Gravity of the object |
| $\ddot{x}_d$ | Desired object acceleration |
| $g$ | Gravitational acceleration |
| $\varphi_d$ | Desired tray orientation |
| $\hat{n}$ | Unit rotation axis |
| $\alpha$ | Rotation angle around $\hat{n}$ |
| $T$ | Moment applied to the object |
| $I$ | Rigid body inertia tensor in CoM coordinate frames |
| $\Omega$ | Object angular velocity |
| $\dot{\Omega}$ | Object angular acceleration |
| $F_c$ | Contact forces |
| $d$ | Diameter of the inscribed circle of the contact surface |
| $\dot{V}_A$ | Acceleration of objects A |
| $\dot{V}_o$ | Acceleration of objects O |
| $P_A$ | Object A's position relative to object O |
| $\Omega_o$ | Angular velocity of object O |
| $\dot{\Omega}_o$ | Angular acceleration of object O |
| $F_A$ | Resultant force exerted on object A |
| $F_{c,A}$ | Contact force exerted on object A |
| $F_{g,A}$ | Gravity exerted on object A |
| $M_A$ | Mass matrix of object A |
| $\mu$ | Friction coefficient |
| $d$ | Side length of the cuboid base of object A |
| $h$ | Height of object A |
| $T_A$ | Moment applied to object A |
| $\Omega_A$ | Angular velocity of object A |
| $\dot{\Omega}_A$ | Angular acceleration of object A |
| $I_A$ | Inertia matrix of object A |
| $P_{zero}$ | ZMP of object A |
| $x_m$ | Operator hand positions acquired via master devices |
| $x_r$ | Reference position |
| $X_r$ | Reference trajectory |
| $\dot{x}_r$ | Reference velocity |
| $\ddot{x}_r$ | Reference acceleration |
| $\boldsymbol{X}_d$ | Desired trajectory smoother states |
| $x_d$ | Desired position |
| $\boldsymbol{u}$ | Desired position fifth-order derivative |
| $\boldsymbol{Q}_d$ | Desired joint trajectory of the robot |
| $q_d$ | Desired robot joint positions |
| $\dot{q}_d$ | Desired robot joint velocities |
| $n$ | Number of manipulator joints |
| $\mathcal{P}_d$ | Desired manipulator end-effector positions and orientations |
| $\mathcal{V}_d$ | Desired end-effector velocities and angular velocities |
| $\varphi_d$ | Desired orientation |
| $q_s$ | Actual robot joint angles |
| $K_{e,x}$ | Position error weight matrix |
| $K_{e,\varphi}$ | Orientation error weight matrix |
| $J$ | Manipulator Jacobian matrix |
| $J^{\dagger}$ | Pseudo-inverse of manipulator Jacobian matrix |
| $\dot{q}_0$ | Manipulator null space motion |
| $\bar{q}_i$ | Upper limit of the $i$-th joint |
| $\underline{q}_i$ | Lower limits of the $i$-th joint |
| $\boldsymbol{X}_s$ | Actual state of the robot |
| $x_s$ | Actual end-effector position |
| $\dot{x}_s$ | Actual end-effector velocity |
| $\ddot{x}_s$ | Actual end-effector acceleration |
| $W_x$ | State weight matrix |
| $W_u$ | Input weight matrix |

**Acknowledgements** This work was supported by the National Natural Science Foundation of China under the Basic Science Center Program for "Space Robot Intelligent Manipulation" (Grant No. T2388101).

**Conflict of Interest** The authors declare that they have no conflict of interest.